\documentclass[11pt]{article}
\usepackage{coling2020}
\usepackage{times}
\usepackage{url}
\usepackage{latexsym}

\usepackage{soul}
\usepackage[utf8]{inputenc}
\usepackage[small]{caption}
\usepackage{rotating, graphicx}
\usepackage{amsmath}
\usepackage{booktabs}
\usepackage{multirow}

\usepackage{xspace}

\newcommand{\Atis}{\mbox{\textsc{ATIS}}\xspace}
\newcommand{\EmailDiag}{\mbox{\textsc{EmailDialogue}}\xspace}
\newcommand{\Sparc}{\mbox{\textsc{SParC}}\xspace}
\newcommand{\Spider}{\mbox{\textsc{Spider}}\xspace}
\newcommand{\Sqa}{\mbox{\textsc{SequentialQA}}\xspace}
\newcommand{\Cosql}{\mbox{\textsc{CoSQL}}\xspace}
\newcommand{\Scone}{\mbox{\textsc{SCONE}}\xspace}
\newcommand{\TimeExp}{\mbox{\textsc{TimeExpression}}\xspace}
\newcommand{\CSQA}{\mbox{\textsc{CSQA}}\xspace}
\newcommand{\SpaceBook}{\mbox{\textsc{SpaceBook}}\xspace}
\newcommand{\TemporalStructure}{\mbox{\textsc{TempStructure}}\xspace}

\newcommand{\SPLASH}{\mbox{\textsc{SPLASH}}\xspace}

\newcommand{\CdSP}{\mbox{\textsc{CdSP}}\xspace}
\newcommand{\CiSP}{\mbox{\textsc{CiSP}}\xspace}

\newcommand{\LFs}{\mbox{\textit{LFs}}\xspace}
\newcommand{\MR}{\mbox{\textit{MR}}\xspace}
\newcommand{\MRs}{\mbox{\textit{MRs}}\xspace}
\newcommand{\Seq}{\mbox{\textsc{Seq2Seq}}\xspace}

\usepackage{amsthm,amsmath,amsfonts,bm,xspace}
\usepackage{color}

\newcommand{\comment}[1]{}

\def\eqref#1{(\ref{#1})}

\def\1{\bm{1}}

\DeclareMathAlphabet{\mathsfit}{\encodingdefault}{\sfdefault}{m}{sl}
\SetMathAlphabet{\mathsfit}{bold}{\encodingdefault}{\sfdefault}{bx}{n}

\colingfinalcopy 

\title{Context Dependent Semantic Parsing: A Survey}

\author{Zhuang Li, Lizhen Qu, Gholamreza Haffari\\
Faculty of Information Technology\\
  Monash University \\
  {\tt firstname.lastname@monash.edu}}

\date{}

\begin{document}

\maketitle

\begin{abstract}
Semantic parsing is the task of translating natural language utterances into machine-readable meaning representations. Currently, most semantic parsing methods are not able to utilize contextual information (e.g. dialogue and comments history), which has a great potential to boost semantic parsing performance. To address this issue, context dependent semantic parsing has recently drawn a lot of attention. In this survey, we investigate progress on the methods for the context dependent semantic parsing, together with the current datasets and tasks. We then point out open problems and challenges for future research in this area. The collected resources for this topic are available at: \url{https://github.com/zhuang-li/Contextual-Semantic-Parsing-Paper-List}.  
\end{abstract}

\section{Introduction}
\blfootnote{
    %
    %
    \hspace{-0.65cm}  
    This work is licensed under a Creative Commons Attribution 4.0 International Licence. Licence details: \url{http://creativecommons.org/licenses/by/4.0/}.
    %
    %
    %
    %
}
Semantic parsing is concerned with mapping natural language (NL) utterances into machine-readable structured \textit{meaning representations} (\MRs). These representations are in the formats of formal languages, e.g. Prolog, SQL, and Python. A formal language is typically defined by means of a formal \textit{grammar}, which consists of a set of rules. Following the convention of the chosen formal language, \MRs are also referred to as logical forms or programs. An \MR is often executable in a (programming) environment to yield a result (e.g. results of SQL queries) enabling automated reasoning~\cite{kamath2018survey}.

Most research work on semantic parsing treats each NL utterance as an \emph{independent} input, ignoring the text surrounding them~\cite{kamath2018survey}, such as interaction histories in dialogues. 
The surrounding text varies significantly across different application scenarios. In a piece of free text, we refer to the surrounding text of a current utterance as its \textit{context}. The context is different with respect to different utterances. In our sequel, we differentiate between context \textit{independent} semantic parsing (\CiSP) and context \textit{dependent} semantic parsing (\CdSP) by whether a corresponding parser utilizes context information. A knowledge base or a database (on which a \MR is executed for the purpose of question answering) can be considered as context as well~\cite{krishnamurthy2012weakly,liang2016learning}.
This type of context does not change with respect to the utterances. In this survey, we only consider the former kind of context which does vary with different utterances.

\begin{table}[ht] 
\centering 
{\small
\begin{tabular}{l l} 
\hline\hline 
$\text{D}$ & Database about pets \\ 
\hline 
$\text{Q}_1$ & What are the different pet types? \\
$\text{S}_1$ & 
SELECT DISTINCT pettype FROM pets\\
$\text{Q}_2$ & For each of those, what is the maximum age? \\
$\text{S}_2$ & SELECT max(pet\_age), pettype FROM pets GROUP BY pettype\\
$\text{Q}_3$ & What about the average age? \\
$\text{S}_3$ & SELECT avg(pet\_age), pettype FROM pets GROUP BY pettype \\
\hline 
\end{tabular}
}
\caption{An example of \CdSP from \Sparc~\cite{yu2019sparc}, where each SQL query $S_i$ is the \MR of the question $Q_i$.}
\label{tab:CoSQL} 
\end{table}
The utilization of context in semantic parsing imposes both challenges and opportunities. 
As shown in Table \ref{tab:CoSQL}, one challenge is to resolve references, such as \textit{those} in ``For each of those, what is the maximum age''. This example shows also another challenge caused by elliptical (incomplete) utterances. The sentence ``What about the average age?'' alone misses information about the database table and the column \textit{pettype}. The incomplete meaning needs to be complemented by the discourse context. 
Compared with \CiSP, which usually assumes that the information within the utterance is complete, \CdSP is expected to tackle challenges posed by involving context in the parsing process  
\cite{liang2016learning,suhr2018learning,zhang2019editing,liu2020FarAwayContextModelingSP}. In addition, tackling the above challenges provides us with more opportunities to inspect the linguistic phenomena which could influence semantic parsing. Our survey on \CdSP fills the gap in the literature, as the recent surveys in the semantic parsing research mainly focus on \CiSP~\cite{kamath2018survey,zhu2019statisticalSPSurvey}. 

This paper is organised as follows. We start with providing a brief and fundamental understanding of  \CiSP in \S2. We then present a comprehensive organization of the recent advances in \CdSP in \S3. We discuss current \CiSP tasks, datasets, and resources in \S4. Finally, we cover  open research problems in \S5, and conclude by providing a roadmap for future research in this area.     

\section{Background}



\CiSP aims to learn a mapping $\pi_{\theta} : \mathcal{X} \rightarrow \mathcal{Y}$, which translates an NL utterance $x \in \mathcal{X}$ into an \MR $y \in \mathcal{Y}$. An \MR $y$ can be executed in a programming environment (e.g. databases, knowledge graphs, etc.) to yield a result $z$, namely denotation. The structure of an \MR takes a form of either a tree or graph, depending on its underlying formal language. The languages of \MRs are categorized into three types of formalism
: logic based (e.g. first order logic), graph based (e.g. AMR~\cite{banarescu2013abstract}), and programming languages (e.g. Java, Python)~\cite{kamath2018survey}. Some semantic parsers explicitly apply a production grammar to yield \MRs from utterances. Such a grammar consists of a set of production rules, which define a list of candidate derivations for each NL utterance. Each derivation deterministically produces a grammatically valid \MR. 

\subsection{Semantic Parsing Models}
Given an utterance $x \in \mathcal{X}$ and its paired \MR $y \in \mathcal{Y}$, a \CiSP model can form a \textit{conditional} distribution $p(y | x)$.
The model learning can be supervised by either utterance-\MR pairs or merely utterance-denotation pairs. If only denotations are available, a widely used approach~\cite{kamath2018survey} is to  marginalize over all possible \MRs for a denotation $z$, which leads to a \textit{marginal} distribution $p(z | x) = \sum_{y} p(z, y| x)$. A parsing algorithm aims to find the optimal \MR in the combinatorially large search space.
We coarsely categorize the existing models into: symbolic approaches, neural approaches, and neural-symbolic approaches based on the category of machine learning methodology and whether any production grammars are explicitly used in models.
\paragraph{Symbolic Approaches}
\label{sec:pipe}
A symbolic semantic parser employs production grammars to generate candidate derivations and find the most probable one via a scoring model. The scoring model is a statistical or machine learning model. 
 Each derivation is represented by handcrafted features extracted from utterances or partial \MRs. Let $\Phi(x, d)$ denote the features of a pair of utterance and derivation, and $G(x)$ be the set of candidate derivations based on $x$. A widely used scoring model is the log linear model~\cite{zettlemoyer2012graphPCCG,kamath2018survey}. 
\begin{equation}
    p( d | x) = \frac{\exp(\bm{\theta}\Phi(x, d))}{\sum_{d' \in G(x)} \exp(\bm{\theta}\Phi(x, d'))}
\end{equation}
where $\bm{\theta}$ denotes the model parameters. If only utterance-denotation pairs are provided at training time, a model marginalizes over all possible derivations yielding the same denotations by $p(z | x) = \sum_{d} p(z, d| x)$~\cite{krishnamurthy2012weakly,liang2016learning}. Those corresponding parsers further differentiate between graph-based parsers~\cite{flanigan2014graphSP,zettlemoyer2012graphPCCG} and shift-reduce parsers~\cite{zhao2014shiftReduceSP} due to the adopted parsing algorithms and the ways to generate derivations. From a machine learning perspective, these approaches are also linked to a structured prediction problem.


\paragraph{Neural Approaches}
\label{sec:end}
Neural approaches apply neural networks to translate NL utterances into \MRs without using production grammars. These approaches formulate semantic parsing as a machine translation problem by viewing NL as the source language and the formal language of \MRs as the target language. 

Most work in this category adopts \Seq \cite{sutskever2014sequence} as the backbone architecture, which consists of an encoder and a decoder. The encoder projects NL utterances into hidden representations, whereas the decoder generates linearized \MRs sequentially.
 Both encoders and decoders employ either recurrent neural networks (RNN)~\cite{goodfellow2016deepbook} or Transformers \cite{vaswani2017attention}. Note that, these methods do not apply any production grammars to filter out syntactically invalid \MRs.

The variants of the \Seq based models also explore structural information of \MRs. \textsc{Seq2Tree} \cite{dong2016language} utilizes a tree-structured RNN as the decoder, which constrains generated \MRs to take syntactically valid tree structures. The \textsc{Coarse2Fine} model~\cite{dong2018coarse} adopts a two-stage generation for the task. In the first stage, a \Seq model is applied to generate \MR templates, which replace entities in \MRs by slot variables for a high-level generalization. In the second stage, another \Seq model is applied to fill the slot variables with the corresponding entities. 
\paragraph{Neural-Symbolic Approaches}
In order to ensure the generated \MRs to be syntactically valid without compromising the generalization power of neural networks, neural-symbolic approaches fuse both symbolic and neural approaches by applying production grammars to the generated \MRs; then the derivations are scored by neural networks. 

The majority of these methods linearize derivations such that they are able to leverage \Seq~\cite{liang2016neural,yin2018tranx,guo2019towards}. At each time step, the decoder of these methods emits either a parse action or a production rule, leading to a grammatically valid \MR at the end. these works produce derivations by varying grammars. \textsc{NSM} \cite{liang2016neural} uses a subset of Lisp syntax. \textsc{TranX}~\cite{yin2018tranx} defines the grammars in Abstract Syntax Description Language, while \textsc{IRNet} \cite{guo2019towards} considers the context-free grammar of a language called SemQL. 


There are also neural-symbolic approaches adopting neural architectures other than \Seq. One of such examples is \cite{andreas2016learning}, which adopts a dynamic neural module network (DNMN) to generate \MRs.



\subsection{Evaluation}
In semantic parsing, \textit{exact match accuracy} is the most commonly used evaluation metric. With \textit{exact match accuracy}, the parsing results are considered correct only when the output \MR/denotations exactly match the string of the ground truth \MR/denotations. One flaw of the evaluation metric is that some types of MRs (e.g., SQL) do not hold order constraints. \newcite{yu2018spider} proposed a metric \textit{set match accuracy} to evaluate the semantic parsing performance over SQLs, which treats each SQL statement as a set of clauses and ignore their orders.

Due to the variety of domains and languages over different datasets, it is difficult to measure all semantic parsing methods in a unified framework. To address this issue, \newcite{yu2018spider}, \newcite{yu2019sparc} and \newcite{yu2019cosql} built different shared-task platforms with leaderboard for semantic parsing evaluation on the common datasets and consistent evaluation metrics.

\section{Context Dependent Semantic Parsing}
Context dependent semantic parsing involves modelling of context in the parsing process. For each current NL utterance, we define its \textit{context} as the information beyond this utterance. With this definition, there are two types of context for semantic parsing, \textit{local} context and \textit{global} context. The \textit{local} context for an utterance is the text and multimedia content surrounding it, which is meaningful only for this utterance. In plain texts, the concept of local context is also quite close to discourse, which is defined as a group of collocated, structured, coherent sentences~\cite{parsing2009speech}. In contrast, its \textit{global} context is the information accessible to more than one utterance, including databases and external text corpora, images or class environment~\cite{iyer2018mapping}. The content of local context varies for each NL utterance while the global context is always static. The work in our survey is only concerned with local context. Therefore, we always refer to "local context" as "context" in the following sections.  

Context provides additional information to resolve ambiguity and vagueness in current utterances. For semantic parsing, one type of ambiguity is caused by references in current utterances, which need to be resolved to previously mentioned objects and relations. References may include explicit or implicit lexical triggers, such as \textit{those} in "For each of those, ..." in our introductory example (Table \ref{tab:CoSQL}). Another ambiguity illustrated by the same example is resulted by ellipsis. The previous context provides constraints to restrict the scope of possible \MRs indicated by current utterances. In addition, context provides information to disambiguate word senses and entities, and link them to knowledge bases to enable complex reasoning. However, semantic parsing literature largely neglects word sense disambiguation, which is regarded as an AI complete problem~\cite{navigli2009wsd:survey}. Last but not least, context allows to exploit discourse coherence for semantic parsing. Coherence relations characterize structural relationships between sentences, thus limit the search space of parse candidates for the following utterances of current ones.

Formally, a context dependent parser takes both an input utterance $x_i$ and its context $C_i$, where $C_i$ could include a broad range of multimedia content. And we consider a group of inter-related utterances with the union set of their context as one \textit{interaction}, $I = (\mathbf{x}, \mathbf{C})$, where $\mathbf{x} = [x_1, ...,x_i,...,x_T]$ and $\mathbf{C} = \cup_{i=1}^{T}C_{i}$. Currently, most \CdSP work focus on the research problems of context $C_i$ regarding the history utterances, \MRs, denotations. Such a parser learns a mapping from a current utterance $x_{i}$ to an \MR $y_i$ by $\pi_{\theta}(x_{i}, C_i)$. 

\subsection{Symbolic Approaches}
Existing symbolic approaches formulate \CdSP as a structured prediction problem by including contextual information into their feature models. Their models capture $ p( d_i | x_i, C_i)$ by including context as a condition. Both \newcite{zettlemoyer2009learning} and \newcite{srivastava2017parsing} divide the parsing process into two steps: i) generate initial parses using \CiSP; ii) complete initial parses using contextual information. In contrast, \newcite{long2016simpler} parses a sequence of utterances in one step. In all those work, symbolic features are used to represent contexts. 

In two-step approaches, \newcite{zettlemoyer2009learning} and \newcite{srivastava2017parsing} differ in the details of individual steps. In the first step, \newcite{zettlemoyer2009learning} extends \MRs with predicates representing references, while \newcite{srivastava2017parsing} generates a set of context independent parses for each utterance. In the second step, \newcite{zettlemoyer2009learning} collects possible derivations by applying three heuristic rules to replace references with entities in context and extend initial \LFs with constraints, then finds the best derivations according to a linear model. In~\cite{srivastava2017parsing}, their model expands the initial parse set with parses selected from context using heuristic rules, then finds the best parses in the expanded set. Their feature model includes a multinomial random variable indicating the current hidden state of discourse.

The shift-reduce parser in~\cite{long2016simpler} generates derivations for a whole utterance sequence. This method stores the previously generated derivations in a stack, performs a sequence of \textit{shift} and \textit{build} operations to generate \LFs. In its feature model, a context is represented by a sequence of past \LFs and a random variable denoting the current world state.

Utterances and \MRs histories form a context of a \CdSP parser. The common practice is to extract handcrafted features from both utterances and \MRs to represent contexts. Some typical feature patterns are as follows: 
\paragraph{Utterance} In~\cite{srivastava2017parsing}, they consider indicator features of lexical triggers, whether the current utterance is repeated, as well as the position of the current utterance in an interaction.  
\paragraph{Meaning Representations} In \cite{zettlemoyer2009learning}, there is a feature indicating if the predicates exist in the history \LFs. Such feature allows the model to learn to copy the segments from the context that contains the expected predicates. \newcite{long2016simpler} adopts the feature indicating if the argument in current \MR is one of arguments in the last \MR. \newcite{srivastava2017parsing} uses the combinations of predicates in successive turns as the indicator features.
\subsection{Neural Approaches}
\begin{figure}
    \centering
    \includegraphics[width=1\textwidth]{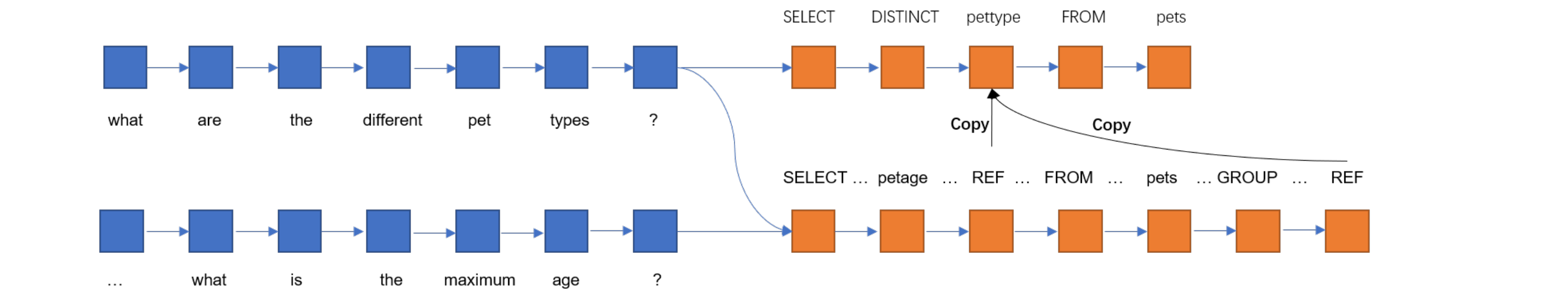}
    \label{fig:semantic_tree}
    \caption{Coreference resolution architecture of \cite{chen2019context}. Considering the example in Table \ref{tab:CoSQL}, \newcite{chen2019context} firstly generates a \MR template for $Q_2$ as "SELECT max(petage), \textit{REF} FROM pets GROUP BY \textit{REF}". The \textit{REF} tokens would then be replaced with the "pettype" from the precedent \MR. }
    \label{fig:sql_example}
\end{figure}
Existing neural \CdSP methods extend the \Seq architecture to incorporate contextual information in two ways. The first approach is to build context-aware encoders to encode historical utterances or \MRs into neural representations, which provide decoders contextual information to resolve ambiguity in current utterances. As previously predicted \MRs provide the constraints and information missed in current utterances, the second approach is to utilize context-aware decoders to reuse or revise those predicted \MRs for generating current \MRs.    

\paragraph{Context-aware Encoders} Encoders of \CdSP methods differentiate between utterance encoders and \MR encoders. Utterance encoders construct neural representations for both current and historical utterances, while \MR encoders build neural representations based on on historical \MRs.

Utterance encoders aim to embed rich information hidden in utterances into fixed-length representations, which provide contextual information in addition to current utterances for decoders. They apply first an RNN to map each utterance into a continuous vector of fixed-size. Then there are three ways to encode utterances in context into a fixed-size neural representation.
\begin{itemize}
    \item 
    For each utterance in a dialogue, a straightforward method is to concatenate its previous $k - 1$ utterances with current utterance in order and encode them with the RNN~\cite{suhr2018learning,suhr2018situated}.
    As a result, decoders have access to information in at most $k$ utterances. However, this method fails to access information beyond the $k$ utterances. In addition, it is computationally expensive because if an utterance belongs to multiple contexts, it would be repeatedly encoded for modelling all the contexts.
 
    \item To overcome the above weakness, an alternative method is to treat a sub-sequence of utterances up to time $t$ as a sequence of vectors, and project them into a \textit{discourse state} vector by using a turn-level RNN~\cite{suhr2018learning,zhang2019editing,he2019pointer}. In another word, those models apply hierarchical RNNs to map each context into a fixed-size vector. In this method, each utterance is encoded only once and reused for modelling different contexts. However, this approach often leads to significant information loss~\cite{pascanu2013difficultyRNN,khandelwal2018sharpRNNContext} due to the challenges imposed by encoding sequences of utterances into single vectors. 
    \item In order to focus on history utterances most relevant to current decoder states or utterances, soft attention~\cite{bahdanau2014neural} is applied to construct context vectors. The query vectors are either the hidden state of an decoder~\cite{suhr2018learning,he2019pointer,suhr2018situated} or an utterance vector~\cite{liu2020FarAwayContextModelingSP,zhang2019editing}. To differentiate between positional information, token embeddings of history utterance are concatenated with their position embeddings~\cite{suhr2018learning,he2019pointer}, which encode the positions of history utterances relative to the current utterances. This method reflects the observation that similar utterances tend to share relevant information, such as references of the same entities. Both discourse states and attended representations are also widely used by the neural dialogue models~\cite{zhang2018context}, thus suffer from the same problems caused by composition complexity. As a result, the trained models are found insensitive to utterance order and word order in context~\cite{sankar2019neuralDialogues}.
\end{itemize}

\MR encoders construct a neural context representation at time $t$ based on the \MRs predicted before $t$. As \MRs are expressed in a formal language, \MR encoders also apply RNNs to encode each \MR or segments of \MRs into embedding vectors. Then \MR encoders build context representations of historical \MRs in the same spirit as utterance encoders. In~\cite{guu2017language}, they only concatenate the embeddings of $k$ most recent history \MR tokens as they assume current \MR is always an extension of previous \MRs. In~\cite{suhr2018learning}, a bidirectional RNN is applied to construct a vector for each segment, which is extracted from historical \MRs. Soft attention is also applied in~\cite{zhang2019editing} for building context vectors, which uses the current hidden state of their decoder as the query vector to attend over the token embeddings of the previous \MR. 

\paragraph{Context-aware Decoders} Decoders in \CdSP models produce \MRs based on the neural representations provided by their encoders. Such a decoder yields an \MR by generating a sequence of \MR tokens according to model distribution $P(\mathbf{y}| \mathbf{x}, \mathbf{C})$, where $\mathbf{C}$ denotes context information. There are three major ways to utilize context information.

One key problem of \CdSP is incomplete information in current utterances. The straightforward way is to take neural context representations $\mathbf{C}$ as additional input of decoders, which are yielded by context-aware encoders. Those context representations contains information from previous utterances, historical \MRs, or both. The decoders take them as input by concatenating them with the ones from current utterances at each decoding step~\cite{suhr2018learning,liu2020FarAwayContextModelingSP,chen2019context,zhang2019editing,shen2019multi}. Thus, the quality of decoding depends tightly on the quality of contextual encoding, which is still a challenging problem~\cite{sankar2019neuralDialogues}.

\MRs of current utterances often contain segments from previous \MRs~\cite{suhr2018learning}. The shared parts are references to previously mentioned entities or constraints implied by context. Reuse of \MR segments is realized by a designated \textit{copy} component, which selects a segment to copy when the probability of copying is high. As decoders in \Seq produce a sequence of decisions for each input, the corresponding model generates a sequence of mixed decisions, including both \textit{copy} of segments and generation of new \MR tokens. In a similar manner,  copying of \MR tokens from previous \MR is proposed in~\cite{zhang2019editing}.

Coreference resolution is explicitly addressed in~\newcite{chen2019context}. As illustrated by the example in Figure \ref{fig:sql_example}, a special token \textit{REF} is introduced in the output vocabulary for denoting if an entity in the preceding \MR is referred in that utterance. If that is the case, the corresponding entity token is copied from the previous \MR to replace the \textit{REF} token via a pointer network module~\cite{vinyals2015pointer}. 

\subsection{Neural-Symbolic Approaches}
\begin{figure}
    \centering
    \includegraphics[width=1\textwidth]{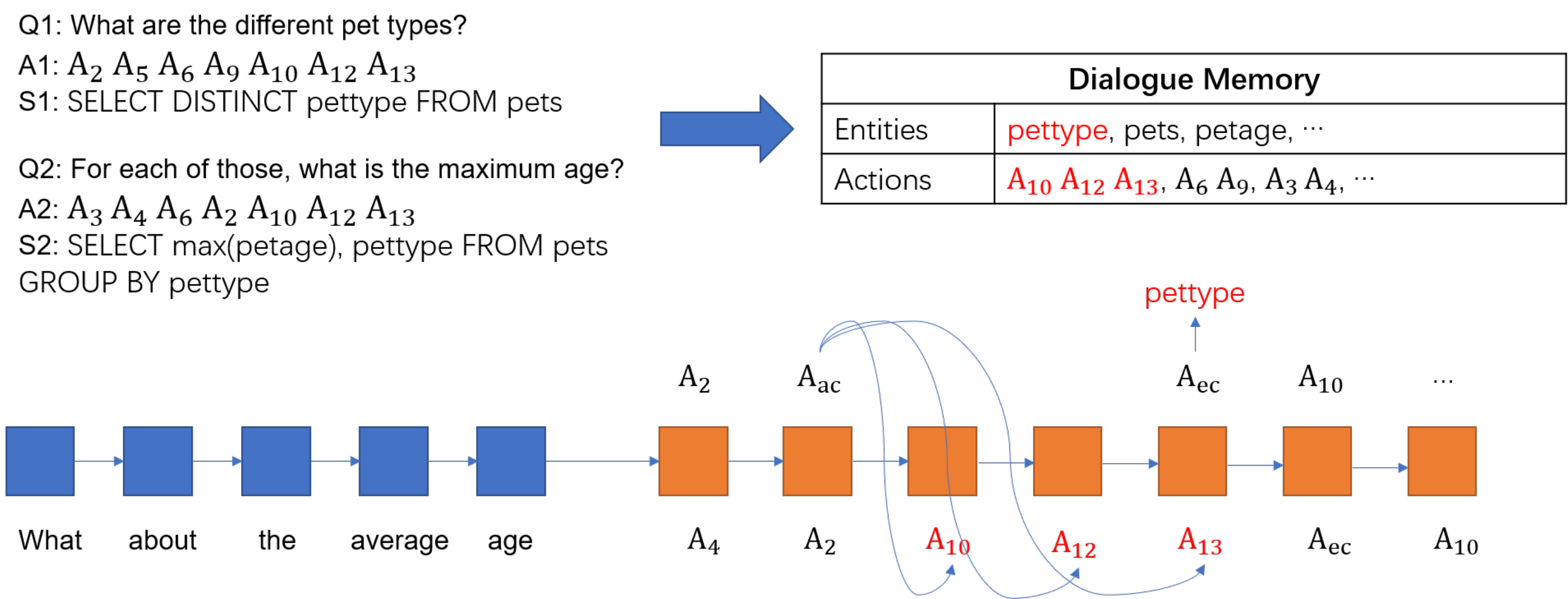}
    \caption{The symbolic memory architecture of \cite{guo2018dialog}. Considering the example in Table \ref{tab:CoSQL}, \newcite{guo2018dialog} defines different types of actions, $A_{ac}$ and $A_{ec}$, to copy action sequence $A_{10}, A_{12}, A_{13}$ and the entity \textit{pettype} from the symbolic memory, respectively.}
    \label{fig:copy_action}
\end{figure}
Neural-symbolic approaches introduce grammar into the decoding process or utilize symbolic representations as intermediate representations, while applying neural nets for representation learning. They take advantages from both the good context representation obtained by neural nets and reduced complexity of decoding due to the constraints introduced by grammars. In existing work, those approaches regard the generation of an \MR as the prediction of a sequence of actions. Neural-symbolic methods normally take the same methods as the neural approaches to encode the contextual information. What differentiate them is the neural-symbolic could handle context by i) designing specific actions, and ii) utilizing symbolic context representations.

The context specific actions proposed in~\cite{iyyer2017search,sun2019knowledge,liu2020FarAwayContextModelingSP} adopt \textit{copy} mechanism to reuse the previous \MRs. \textsc{CAMP}~\cite{sun2019knowledge} include three actions to copy three different SQL clauses from precedent queries. 
\newcite{liu2020FarAwayContextModelingSP} allows copying of any actions or subtrees from precedent SQL queries. The \textit{subsequent} action in~\cite{iyyer2017search} adds SQL conditions from the previous query into the current semantic parse to address the ellipsis problem. Different from other approaches, \newcite{iyyer2017search} uses a \textsc{DynSP}, which is in a similar neural network structure as the DNMN, instead of the \Seq to generate the action sequences.

Production rules are also used to explicitly address the coreference resolution. In~\cite{shen2019multi}, the authors defined fours actions to instantiate the entities, predicates, types and numbers. Then the pointer network is utilized to find mentions of the four entry semantic categories in the current and history utterances. The entities in utterances are later mapped to entities in knowledge bases by using their entity linking tool.

Instead of directly copying from previous \MRs, the parser \textsc{Dialog2Action}~\cite{guo2018dialog} incorporates a dialogue memory, which maintains \textit{symbolic} representations of entities, predicates and action subsequences from an interaction history (Figure \ref{fig:copy_action}). That parser defines three types of designated actions to copy entities, predicates and action subsequences from the memory respectively. Instead of decisively copying from memory, each type of action probabilistically selects the corresponding segments conditioning on the \textit{symbolic} representations, which are later integrated into the generated action sequences. 

\newcite{guo2019coupling} employs the same neural-symbolic models as in \cite{guo2018dialog} to capture contextual information. Different from other approaches, \newcite{guo2019coupling} adopts the meta-learning approach to improve the generation ability of \CdSP models. Inspired by \cite{huang2018natural}, \newcite{guo2019coupling} utilize the context from other interactions to guide the learning of \CdSP over utterances within current interactions via MAML. \newcite{guo2019coupling} considers an input utterance $x_i$ and its context $C_i$ as an instance. A context-aware retriever would retrieve instances, which are semantically close to the current instances, from other interactions. When learning model parameters, the retrieved instances and the current instances are considered as the support set and test set, respectively, and grouped as tasks as in the common MAML paradigm.

\subsection{Comparison between Different \CdSP Approaches}
In~\cite{liu2020FarAwayContextModelingSP}, 13 different context modeling methods for both neural and neural-symbolic \CdSP parsers were evaluated on two benchmark datasets. None of those methods achieve consistent superior results over the others in all experimental settings. Among them, concatenation of $k$ recent utterances for decoders and copy of parse actions from precedent \MRs are the top performing ones in most settings. ~\newcite{liu2020FarAwayContextModelingSP} defines 12 fined-grained types summarized with multiple hierarchies according to the contextual linguistic phenomena, and inspects how different linguistic phenomena influence the model behavior. One interesting conclusion is that the methods in their experiments all perform poorly on the instances involving coreference problems that require complex inference. But note that, those methods in that study were not compared with the ones with explicit coreference resolution. Another interesting finding is that all the models perform better on the utterances which only augment the semantics of previous sentences than on the utterances which substitute the partial semantics of the precedent utterances. 

\subsection{Comparison between \CdSP and Feedback Semantic Parsing}
Feedback/Interactive Semantic Parsing is another line of research in semantic parsing that utilizes context to refine \MRs in an iterative manner. Most Feedback Semantic Parsing systems~\cite{iyer2017learning,yao2019model,yao2019interactive,elgohary2020speak} start with using an \CiSP parser to parse a given utterance into an initial \MR. Then the \MR is interpreted in natural language and sent to a user. The user provides feedback, based on which the systems revise the initial parse. The process repeats till convergence. Therefore, in Feedback Semantic Parsing, interaction histories are only used to revise the parses. In contrast, \CdSP focuses on modelling the dependencies between the utterances. \newcite{elgohary2020speak} empirically compares \CdSP with Feedback Semantic Parsing. They train a \CdSP model, EditSQL~\cite{zhang2019editing}, on two \CdSP datasets, \Sparc and \Cosql, and evaluate it on the test set of a feedback semantic parsing dataset, \SPLASH. The performance is merely 3.4\% and 3.2\% in terms of accuracy, indicating that the two tasks are distinct by addressing different aspects of context.

\section{Datasets and Resources}
\begin{table*}[ht] 
\centering 
{
\resizebox{\columnwidth}{!}{\begin{tabular}{l l c c c c c c } 
\hline\hline 
\textbf{Datasets} & \textbf{Reference}& \textbf{\#Party} & \textbf{Annotation} & \textbf{MR Language} & \textbf{\#Utterance} & \textbf{\#Interaction} & \textbf{Avg. \#Turns} \\ 
\hline 
\Atis & \cite{price1990evaluation}& 1 & MR & SQL/lambda  &  11,653 & 1,658 & 7.0 \\ 
\Sqa & \cite{iyyer2017search} & 1 & Denotation & self & 17,553 & 6,066 & 2.9 \\ 
\Sparc & \cite{yu2019sparc} & 1 & MR & SQL & 12,726 & 4,298 & 3.0 \\ 
\TimeExp & \cite{lee2014context} & 1 & Denotation & lambda & NA & 298 & NA \\ 
\TemporalStructure & \cite{chen2019context} & 1 & MR & self & 1,237 & NA & NA \\ 
\Scone & \cite{long2016simpler} & 1 & Denotation & self & 69,755 & 13,951 & 5.0 \\ 
\CSQA & \cite{saha2018complex} & 1 & Denotation & SPARQL/self & \char`\~1.6M & \char`\~200,000 & \char`\~10.0 \\
\hline\hline
\SpaceBook & \cite{vlachos2014new} & 2 & MR, act & self & 2,374 & 17 & 139.7 \\ 
\EmailDiag & \cite{srivastava2017parsing} & 2 & MR & Lisp & 4,759 & 113 & 42.0 \\ 
\Cosql & \cite{yu2019cosql} & 2 & MR, act & SQL & 15,598 & 3,007 & 5.2 \\ 
\hline 
\end{tabular}
}
}
\caption{The statistics of the context dependent datasets. "\textbf{\#}" denotes the number of the corresponding units (e.g. number of utterances, number of interactions, etc.). "\char`\~" denotes this is an estimated number. "NA" denotes that the corresponding statistic data is not applicable. 
"self" denotes the target languages in the datasets are only applicable to a small range of datasets.}
\label{tab:context_data}
\end{table*}
Table \ref{tab:context_data} summarizes the basic properties and statistics of existing \CdSP datasets. There are two scenarios of the \CdSP datasets, Single-party Scenarios and Multi-party Scenarios. In the former scenarios, the user utterances are translated into \MRs to obtain the execution results from the programming environment. In the latter scenarios, there are systems which respond to the users in natural language based on the user utterances and the execution results.
The user utterances are manually labeled with different types of annotations, including \MRs, denotations, and dialogue acts. The system responses are usually annotated with the dialogue acts. We especially highlight those annotations that explicitly reflect contextual dependencies of utterances in the sequel. 
\subsection{Scenarios}
\paragraph{Single-party Scenarios} In \Sparc~\cite{yu2019sparc}, \Sqa~\cite{iyyer2017search} and \Atis, the user utterances within each interaction are around a topic described by the provided text. To collect \Sparc and \Sqa, crowd-workers are asked to raise questions to obtain the information that answers the questions sampled from other corpora~\cite{pasupat-liang-2015-compositional,yu2018spider}. But the assumption for \Sqa is the answers of the current question must be the subset of answers from the last turn. In \Atis~\cite{price1990evaluation}, crowd-workers raise questions around the detailed scripts describing air travel planning scenarios. 

\TemporalStructure~\cite{chen2019context} and \TimeExp~\cite{lee2014context} particularly focused on addressing the temporal-related dependency. In \TemporalStructure, human users or the simulators raise natural language questions chronologically towards a knowledge base. The facts in the knowledge base are organized in time series. Therefore, the questions in \TemporalStructure are rich with time expressions. \TimeExp only annotate temporal mentions (text segments that describe time expressions) instead of complete questions. All the mentions are from the time expression-rich corpora.


\Scone~\cite{long2016simpler} and \CSQA~\cite{saha2018complex} use semi-automatic approaches to simulate the contextual dependency. Each interaction in \Scone is merely labeled with an initial denotation and an end denotation. The denotations in \Scone are regarded as the states that can be manipulated by the programs. Within each interaction, multiple candidate sequences of programs would be automatically generated while only the sequence of programs, which could correctly transit the initial state to the end state, would be kept and described with natural language by the crowd-workers. To create \CSQA \cite{saha2018complex} dataset, the crowd-workers are asked to raise questions that can be answered from single fact tuples (e.g. relation: \textit{CEO}, subject: \textit{Google}, object: \textit{Sundar Pichai}) in the knowledge graph or the complex facts which are the composition of multiple tuples. To create coherent dependency among questions, the questions that share the relations or entities are placed next to each other. And crowd-workers would manually modify the questions such that the sequence of questions would include contextual linguistic properties such as ambiguity, underspecification or coreference. It is worth mentioning that, with such method, \CSQA includes the largest number of interactions until now, which is over 200k.
\paragraph{Multi-party Scenarios}
Similar to the scenario of \Sparc, to obtain the answers to the questions sampled from \Spider~\cite{yu2018spider}, the conversations in \Cosql~\cite{yu2019cosql} are conducted between two human interlocutors, who play the roles of user and system, respectively. The dialogues in the \SpaceBook~\cite{vlachos2014new} are under the scenarios formed by the routing requests. One human interlocutor pretends to be a tourist walking around Edinburgh while another interlocutor plays the role of a system responding to the tourist. The conversations in \EmailDiag are between the human agent and an email assistant instead of two humans. 
\subsection{Context and Annotations}
The contextual linguistic phenomena in the \CdSP corpora is quite close to the phenomena in the corpora of tasks such as document-level machine translation, question answering, dialogue system, etc.. However, in \CdSP datasets, the contextual linguistic phenomena has a tight relation with the annotations.

\newcite{iyyer2017search}, \newcite{vlachos2014new} defined specific components in the \MR languages of \Sqa and \SpaceBook to explicitly model the context dependency. \Sqa introduced a keyword \textit{subsequent}. All the answers of \MR statements after \textit{subsequent} would only be the subset of the answers of the precedent \MR. In the language of \SpaceBook, to resolve the coreference problem, a special predicate \textit{equivalent} could indicate the identical entities across questions at different turns.

The context dependency could be reflected by some properties of annotations. \newcite{yu2019sparc} analyzed semantic changes over turns in \Sparc by calculating the overlapping percentage of tokens between the SQL annotations at different turns. In \Sparc, the average overlapping percentage increases at later turns within one interaction, where the users tend to narrow down their topics with turns increasing. Both \newcite{yu2019sparc} and \newcite{liu2020FarAwayContextModelingSP} categorized the contextual phenomena in \Sparc into fine-grained types and calculate their frequency. \newcite{yu2019sparc} found some SQL representations correspond to certain contextual phenomena types. For instance, in the questions of the \textit{theme-entity}, which means the current question and precedent question are around the same entities but request for different properties, their corresponding SQL representations have the same \textit{FROM} and \textit{WHERE} clauses. But the SQL representations for other types may vary.

For the datasets, \SpaceBook and \Cosql, \newcite{yu2019cosql} and \newcite{vlachos2014new} label utterances with dialogue acts along with \MRs. Different from \cite{yu2019cosql}, \newcite{vlachos2014new} integrated the dialogue acts into the \MRs. The dialogue acts can be considered as the overall functions of the utterances while different dialogue acts reflect different properties of utterances. For example, in \Cosql, the unanswerable questions that can not be parsed into SQLs are labelled with dialogue acts such as \textit{NOT\_RELATED}, \textit{CANNOT\_UNDERSTAND}, or \textit{CANNOT\_ANSWER}. The ambiguous questions that need to be clarified are labelled with \textit{AMBIGUOUS}. The following questions are then labeled with \textit{CLARIFY}. The dialogue acts could provide additional contextual information for \CdSP.


\section{Challenges and Future Directions}
\CdSP distinct from \CiSP by context modelling and utilization of context information in the parsing process to complete missing information in \MRs. Despite significant progress in recent years, there are still multiple directions worth pursuing. 
\paragraph{Analysis of Linguistic Phenomena Benefiting from Context} \newcite{yu2018spider} and \newcite{liu2020FarAwayContextModelingSP} analyzed the influence of different types of contextual information on \CdSP methods. Despite some empirical results, it still lacks of a thorough understanding of pros and cons of each type of context in relation to the parsing task. For example, in which cases should parsers extract information from \MRs in context instead of utterances? Apart from ellipsis and coreference resolution, are there other linguistically motivated problems in context the current parsers have not addressed yet? 
\paragraph{Incorporating Far-side Pragmatics} Current \CdSP approaches fall in the scope of near-side pragmatics, in particular reference resolution, and current \CdSP datasets (e.g. \Cosql, \SpaceBook) consider dialogue acts as merely the overall function of the utterances~\cite{vlachos2014new}. However, \textit{far-side pragmatics} focuses on what happens beyond saying, including implicatures and communicative intentions etc.. Incorporating far-side pragmatics in semantic parsing will be especially useful towards completely understanding dialogues. Thus, there is a need to create large corpora annotated with rich information about various aspects of pragmatics for both training and evaluation.

\paragraph{Causal Structure Discovery in Context} A key challenge of context based modelling is composition complexity caused by highly varying context. The empirical results in~\cite{liu2020FarAwayContextModelingSP} show that the SOTA models can capture well nearby context information but it is still challenging to capture long-range dependencies in context. One possible direction is to find out the underlying causal structure~\cite{glymour2014discovering}, which should be sparse and explains well which contextual information leads to current utterances. If we can focus only on the key reasons in context that lead to changes of \MRs, the influence from noisy information and overfitting of models is expected to decrease significantly. Another potential benefit of understanding causal structures in context is to improve robustness of parsers by ignoring non-robust features~\cite{zhang2020adversarial,ilyas2019adversarial}.

\paragraph{Low-resource \CdSP} Since most \CdSP datasets are small in terms of the number of utterances and interactions, the direction on addressing the low-resource problem in \CdSP is quite promising. The meta-learning approaches, such as the MAML \CdSP in \cite{guo2019coupling}, could be a potential direction to address this issue. The other typical methods to solve low-resource issues, including weakly supervision, data augmentation, semi-supervised learning, self-supervised learning etc., could be further investigated in the scenarios of \CdSP.

\section*{Acknowledgements}
We thank Xuanli He and anonymous reviewers for their useful suggestions. 
\bibliographystyle{coling}
\bibliography{coling2020}

\end{document}